\documentclass[runningheads,orivec]{llncs}
\usepackage[T1]{fontenc}
\usepackage{graphicx}
\usepackage{microtype}
\usepackage{booktabs}
\usepackage{multirow}
\usepackage{amsmath}
\usepackage{arydshln}
\usepackage{adjustbox}
\usepackage{todonotes}
\usepackage{color,xcolor,colortbl}
\usepackage[framemethod=TikZ]{mdframed}
\usepackage{pgfplots}
\usepgfplotslibrary{fillbetween}
\usetikzlibrary{patterns}
\usepackage{subcaption}

\pgfplotsset{compat=1.18}

\newcommand{\quotes}[1]{``#1''}
\renewcommand{\vec}[1]{\ensuremath{\mathbf{#1}}}

\makeatletter
\def\adl@drawiv#1#2#3{%
        \hskip.5\tabcolsep
        \xleaders#3{#2.5\@tempdimb #1{1}#2.5\@tempdimb}%
                #2\z@ plus1fil minus1fil\relax
        \hskip.5\tabcolsep}
\newcommand{\cdashlinelr}[1]{%
  \noalign{\vskip\aboverulesep
           \global\let\@dashdrawstore\adl@draw
           \global\let\adl@draw\adl@drawiv}
  \cdashline{#1}
  \noalign{\global\let\adl@draw\@dashdrawstore
           \vskip\belowrulesep}}
\makeatother

\definecolor{lightblue}{HTML}{5DA5DA}
\definecolor{lightorange}{HTML}{FAA43A}
\definecolor{lightgreen}{HTML}{60BD68}
\definecolor{lightpurple}{HTML}{B276B2}
\definecolor{brown}{HTML}{D95F0E}
\definecolor{lightgrey}{HTML}{BDBDBD}

\title{Few-shot Cross-lingual Aspect-Based Sentiment Analysis with Sequence-to-Sequence Models}

\titlerunning{Few-shot Cross-lingual Aspect-Based Sentiment Analysis}

\author{Jakub \v{S}m\'{i}d\inst{1,2}\orcidID{0000-0002-4492-5481} \and Pavel P\v{r}ib\'{a}\v{n}\inst{1}\orcidID{0000-0002-8744-8726} \and Pavel Kr\'{a}l\inst{1,2}\orcidID{0000-0002-3096-675X}}
 \institute{University of West Bohemia in Pilsen\\
          Faculty of Applied Sciences, Department of Computer Science and Engineering\\
          \and
          NTIS -- New Technologies for the Information Society\\
          Univerzitni 27328, 301 00 Plze\v{n}, Czech Republic\\
          \email{\{jaksmid,pribanp,pkral\}@kiv.zcu.cz} \\
            \tt {\url{https://nlp.kiv.zcu.cz}} \\
          }

\begin{document}
\maketitle
\begin{abstract}
Aspect-based sentiment analysis (ABSA) has received substantial attention in English, yet challenges remain for low-resource languages due to the scarcity of labelled data. Current cross-lingual ABSA approaches often rely on external translation tools and overlook the potential benefits of incorporating a small number of target language examples into training. In this paper, we evaluate the effect of adding few-shot target language examples to the training set across four ABSA tasks, six target languages, and two sequence-to-sequence models. We show that adding as few as ten target language examples significantly improves performance over zero-shot settings and achieves a similar effect to constrained decoding in reducing prediction errors. Furthermore, we demonstrate that combining 1,000 target language examples with English data can even surpass monolingual baselines. These findings offer practical insights for improving cross-lingual ABSA in low-resource and domain-specific settings, as obtaining ten high-quality annotated examples is both feasible and highly effective.
\keywords{Cross-lingual aspect-based sentiment analysis \and Aspect-based sentiment analysis  \and Sentiment analysis \and Transformers}
\end{abstract}

\section{Introduction}

Aspect-based sentiment analysis (ABSA) is a fine-grained sentiment analysis task that goes beyond assigning an overall sentiment label to a piece of text. Instead, it identifies specific opinion targets -- such as products, services, or their attributes -- and determines the sentiment expressed toward each of them. To capture this information, ABSA typically models three sentiment elements~\cite{SMID2025103073}: aspect term ($a$), aspect category ($c$), and sentiment polarity ($p$). In the sentence \textit{\quotes{The staff was helpful}}, the elements correspond to \textit{\quotes{staff}}, \textit{\quotes{service}}, and \textit{\quotes{positive}}, respectively. Some inputs include implicit aspect terms, as in \textit{\quotes{Delightful experience!}}, where the target is not stated directly and is often labelled as \textit{\quotes{NULL}}.

ABSA tasks vary in complexity based on predicted elements and joint extraction. Early work focused on simpler tasks like aspect term extraction and sentiment polarity classification. Recent work has shifted towards compound tasks that jointly predict multiple sentiment elements, such as aspect category sentiment analysis (ACSA)~\cite{schmitt-etal-2018-joint}, end-to-end ABSA (E2E-ABSA)~\cite{wang2018towards}, aspect category term extraction (ACTE)~\cite{pontiki-etal-2016-semeval}, and target-aspect-sentiment detection (TASD)~\cite{tasd}. Table~\ref{tab:absa-tasks} presents selected ABSA task output formats.

\begin{table}[ht!]
    \centering
    \caption{Outputs of selected ABSA tasks for input: \textit{\quotes{Great soup, expensive coffee}}.}
    \begin{adjustbox}{width=0.67\linewidth}
        \begin{tabular}{@{}lll@{}}
            \toprule
            \textbf{Task} &  \textbf{Output}     & \textbf{Example output}  \\                        \midrule
            ACSA      &  \{($c$, $p$)\}      & \{(food, POS), (drinks, NEG)\}               \\
            E2E-ABSA      &  \{($a$, $p$)\}      & \{(\quotes{soup}, POS), (\quotes{coffee}, NEG)\}               \\
            ACTE          &  \{($a$, $c$)\}      & \{(\quotes{soup}, food), (\quotes{coffee}, drinks)\}           \\
            TASD          & \{($a$, $c$, $p$)\} & \{(\quotes{soup}, food, POS), (\quotes{coffee}, drinks, NEG)\} \\ \bottomrule
        \end{tabular}
    \end{adjustbox}
	\label{tab:absa-tasks}
\end{table}

Although ABSA has been widely studied in English, multilingual support is essential for real-world applications, where annotating target-language data is costly. Cross-lingual ABSA transfers knowledge from high-resource languages like English to low-resource ones. Recent work~\cite{li2020unsupervised,lin2023clxabsa,zhang-etal-2021-cross} combines multilingual models like XLM-R~\cite{conneau-etal-2020-unsupervised} with machine translation to generate pseudo-labelled data, but this often fails to capture language-specific nuances in user-generated texts. It is also less effective for complex tasks like TASD, which require structured predictions. Sequence-to-sequence models with constrained decoding offer a more effective alternative~\cite{icaart25}.

While most cross-lingual ABSA studies focus on zero-shot transfer, few-shot examples -- such as a small number of high-quality target language annotations -- can be much more resource-efficient. However, the impact of such examples remains underexplored. In this work, we address these gaps by investigating the effects of adding a limited number of labelled target-language examples on compound cross-lingual ABSA tasks.

Our main contributions include: 1) A comprehensive evaluation of four ABSA tasks across six target languages using two state-of-the-art sequence-to-sequence models, including a previously unexplored task and target language in cross-lingual ABSA. 2) Demonstrating that even ten target language examples lead to substantial performance improvements, reducing the need for techniques like constrained decoding. 3) Showing that few-shot models can outperform monolingual baselines with sufficient target language supervision. 4) An error analysis highlighting the most challenging elements of sentiment prediction, offering insights for future improvements.

\section{Related Work}
Early cross-lingual ABSA research~\cite{barnes-etal-2016-exploring,jebbara-cimiano-2019-zero,lambert-2015-aspect} focused on simple tasks involving a single sentiment element using machine translation or cross-lingual embeddings. More recent work~\cite{li2020unsupervised,lin2023clxabsa,LIN2024125059,zhang-etal-2021-cross} targets E2E-ABSA with machine translation and multilingual encoder-only Transformer models like XLM-R~\cite{conneau-etal-2020-unsupervised}, enhanced by techniques like distillation~\cite{zhang-etal-2021-cross}, contrastive learning~\cite{lin2023clxabsa}, or dynamic loss weighting~\cite{LIN2024125059}.

Sequence-to-sequence models like mT5~\cite{xue-etal-2021-mt5} and mBART~\cite{tang2020multilingual} have been applied to cross-lingual ABSA with constrained decoding as an alternative to machine translation~\cite{icaart25}. These models handle compound tasks such as E2E-ABSA, ACTE, and TASD, demonstrating flexibility in capturing complex sentiment structures. Fine-tuned large language models are also explored~\cite{icaart25}, extending earlier work in monolingual English ABSA~\cite{smid-etal-2024-llama}.

\section{Methodology}
\begin{figure*}[ht!]
    \centering
    \includegraphics[width=0.97\linewidth]{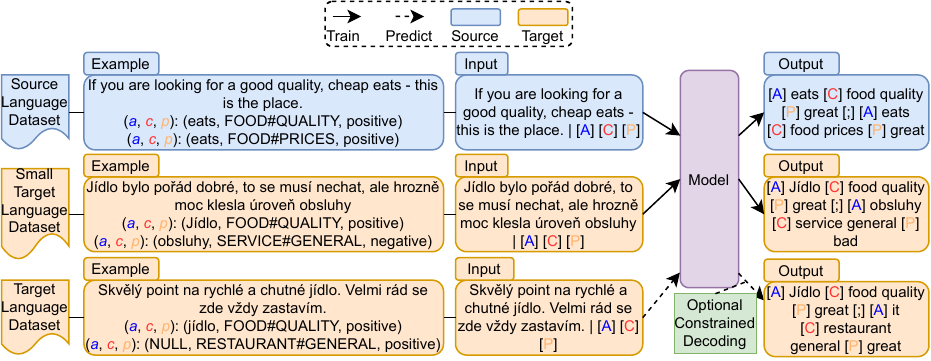}
    \caption{Overview of the training approach. The method involves converting input labels into natural language phrases, fine-tuning on source language data alongside a few target language examples, and generating predictions on target language inputs. Constrained decoding is optionally used to improve output quality.}
    \label{fig:overview}
\end{figure*}

In this section, we describe our method for tackling the triplet task (TASD), which can be readily adapted to tuple-based tasks with minimal modifications. Figure~\ref{fig:overview} provides an overview of the proposed approach.

\subsection{Problem Definition}
Given an input sentence, the goal is to predict all sentiment tuples $T={(a, c, p)}$, where each tuple consists of an aspect term ($a$), aspect category ($c$), and sentiment polarity ($p$). Following previous work~\cite{icaart25} for a fair comparison, we convert each element ($a, c, p$) into a natural language representation ($e_a, e_c, e_p$), with a few examples illustrated in Figure~\ref{fig:overview}. For example, we map implicit aspect terms to \textit{\quotes{it}}, \textit{\quotes{positive}} sentiment polarity to \textit{\quotes{great}}, and \textit{\quotes{negative}} sentiment polarity to \textit{\quotes{bad}}, while keeping explicit aspect terms and aspect categories in their original form.

\subsection{Input and Output Construction}

To build input and output sequences for the model, we follow the strategy proposed in~\cite{icaart25}, using special markers to denote each sentiment element: {\sffamily[A]} for $e_a$, {\sffamily[C]} for $e_c$, and {\sffamily[P]} for $e_p$. These markers are used to prefix each element in the target sequence and are also appended to the input sequence to guide the model’s decoding process. For instance, given the sentence \textit{\quotes{The staff was very helpful}} yield the output: \textit{\quotes{{\sffamily[A]} staff {\sffamily[C]} service general {\sffamily[P]} great}}. If a sentence contains multiple sentiment tuples, we concatenate their representations using the {\sffamily[;]} separator. Figure~\ref{fig:overview} includes several examples of such input-output pairs.

\subsection{Constrained Decoding}
Constrained decoding (CD)~\cite{constrained}, shown in~\cite{icaart25} to be effective for cross-lingual ABSA with sequence-to-sequence models, mitigates errors where a model trained on a source language (e.g. English) generates aspect terms in the source language instead of target one (e.g. Dutch). CD limits token generation based on the input and prior outputs. For example, when generating an aspect category, only tokens from the set of valid categories are allowed. For aspect terms, generation is limited to tokens from the original sentence and \textit{\quotes{it}} for implicit aspects. We use the same CD method as in~\cite{icaart25} and evaluate whether few-shot target language examples can offer similar benefits, potentially reducing the need for CD.

\subsection{Training}
We fine-tune a pre-trained sequence-to-sequence (encoder-decoder) model on constructed input–output pairs. The encoder processes the input sequence $x$ into a contextual representation $\vec{e}$, and the decoder generates the output $y$ by modelling the probability $P_{\vec{\Theta}}(y|\vec{e})$, where $\vec{\Theta}$ denotes the model parameters. During training, each token $y_i$ is predicted based on $\vec{e}$ and previously generated tokens. The model is optimized by minimizing the negative log-likelihood loss over a target sequence of length $n$:
\begin{equation}
    \mathcal{L} = -\sum_{i=1}^n\log p_{\vec{\Theta}}(y_i|\vec{e}, y_{<i}).
\end{equation}

\section{Experimental Setup}
We conduct experiments on four ABSA tasks: ACSA, E2E-ABSA, ACTE, and TASD. To the best of our knowledge, this paper is the first to examine ACSA in cross-lingual settings.

We conduct experiments using the SemEval-2016 dataset~\cite{pontiki-etal-2016-semeval} with restaurant reviews in six languages: English (en), Spanish (es), French (fr), Dutch (nl), Russian (ru), and Turkish (tr), with official training and test splits. A validation set is created by splitting the original training data in a 9:1 ratio, as in~\cite{icaart25}. Additionally, we use the \texttt{CsRest-M} dataset~\cite{smid-etal-2024-czech}, containing real-world restaurant reviews in Czech (cs), which is divided into training, validation, and test sets. Czech has not been explored in cross-lingual ABSA before. Table~\ref{tab:data_stats} summarizes dataset statistics for each language.

\begin{table}[ht!]
    \centering
   \caption{Dataset statistics for each language. }
    \begin{adjustbox}{width=0.5\linewidth}
        \begin{tabular}{@{}llrrrrrrr@{}}
            \toprule
                                   &           & \textbf{Cs}   & \textbf{En}  & \textbf{Es}   & \textbf{Fr}     & \textbf{Nl} & \textbf{Ru}   & \textbf{Tr} \\ \midrule
            \multirow{2}{*}{Train} & Sentences & 2,151   & 1,800        & 1,863         & 1,559           & 1,549       & 3,289         & 1,108       \\
                                   & Triplets  & 4,386   & 2,266        & 2,455         & 2,276           & 1,676       & 3,697         & 1,386       \\
                                   \cdashlinelr{1-9}
            \multirow{2}{*}{Dev}   & Sentences  & 240   & 200          & 207           & 174             & 173         & 366           & 124         \\
                                   & Triplets   & 483  & 241          & 265           & 254             & 184         & 392           & 149         \\
\cdashlinelr{1-9}
            \multirow{2}{*}{Test}  & Sentences & 798   & 676          & 881           & 694             & 575         & 1,209         & 144         \\
                                   & Triplets  & 1,609   & 859          & 1,072         & 954             & 613         & 1,300         & 159   \\      \bottomrule
        \end{tabular}
    \end{adjustbox}
     
    \label{tab:data_stats}
\end{table}

\subsection{Experimental Details}

We employ multilingual sequence-to-sequence models, mT5~\cite{xue-etal-2021-mt5} and mBART~\cite{tang2020multilingual}, accessed via the HuggingFace Transformers library~\cite{wolf-etal-2020-transformers}. For all experiments, each model is fine-tuned for 20 epochs with a batch size of 16, using greedy decoding during inference. We use Adafactor~\cite{adafactor} optimizer with a learning rate of 1e-4 for mT5 and AdamW~\cite{adamw} with a learning rate of 1e-5 for mBART. Hyperparameters were selected for stable validation performance across languages and tasks. Experiments run on an NVIDIA L40 GPU with 48 GB of memory.

We use English as the source language and the other languages as target languages for all experiments and perform model selection based on performance on the English validation set. For few-shot experiments, we select the first~$n$ examples from the target language’s training set. The data are not ordered by label or difficulty, and the initial subset reflects the overall label distribution, ensuring representative coverage. This approach also ensures consistency across runs and models.

\subsection{Evaluation Metrics \& Compared Methods}
We report results averaged over five runs with different random seeds and 95\% confidence intervals. The primary evaluation metric is the micro F1-score, standard in ABSA research, where a predicted sentiment tuple is considered correct only if all components exactly match the gold standard.

We compare our few-shot cross-lingual results with the zero-shot baselines reported in~\cite{icaart25}, where applicable -- that is, for all languages except Czech and all tasks except ACSA. Zero-shot results for Czech and ACSA, as well as all few-shot results, are newly reported in this paper. Using the same models and data splits ensures a fair evaluation of the few-shot effect.

\section{Results}

Table~\ref{tab:res_mt5} presents the results for the four ABSA tasks using mT5, comparing different numbers of few-shot examples and monolingual training. As expected, monolingual training consistently yields the best performance across all tasks and languages, highlighting the advantage of abundant in-language supervision.

\begin{table}[ht!]
\centering
\caption{Cross-lingual micro F1-scores on four tasks in six target languages using mT5, compared to monolingual (mono) performance, with varying few-shot examples (FS). \textbf{Bold} marks better constrained decoding (CD) results with non-overlapping 95\% confidence intervals. Asterisked (*) results are from~\cite{icaart25}.}
\begin{adjustbox}{width=\linewidth}
\begin{tabular}{@{}ccr@{\hspace{6pt}}rrrrrrrrrrrr@{}}
\toprule
\multirow{2}{*}{\textbf{Task}} & \multirow{2}{*}{\textbf{Setup}} & \multirow{2}{*}{\textbf{FS}} & \multicolumn{6}{c}{\textbf{Without constrained decoding}}                                                 & \multicolumn{6}{c}{\textbf{With constrained decoding}}                                                                                                          \\ \cmidrule(lr){4-9}  \cmidrule(lr){10-15}
                               &                                    &                                    & \multicolumn{1}{c}{Cs} & \multicolumn{1}{c}{Es} & \multicolumn{1}{c}{Fr} & \multicolumn{1}{c}{Nl} & \multicolumn{1}{c}{Ru} & \multicolumn{1}{c}{Tr} & \multicolumn{1}{c}{Cs}   & \multicolumn{1}{c}{Es}   & \multicolumn{1}{c}{Fr}   & \multicolumn{1}{c}{Nl}   & \multicolumn{1}{c}{Ru}   & \multicolumn{1}{c}{Tr}                       \\ \midrule
\multirow{8}{*}{\rotatebox[origin=c]{90}{ACSA}}          & Mono                        &                                    & 76.6$^{\pm0.2}$ & 77.1$^{\pm0.2}$ & 69.2$^{\pm0.8}$ & 74.1$^{\pm0.3}$ & 78.0$^{\pm0.8}$ & 74.2$^{\pm1.5}$ & 76.5$^{\pm1.1}$          & 77.4$^{\pm0.6}$          & 69.0$^{\pm0.7}$          & 74.3$^{\pm0.9}$          & 77.7$^{\pm0.5}$          & 74.4$^{\pm4.6}$          \\  \cdashlinelr{2-15}
                               & \multirow{7}{*}{\rotatebox[origin=c]{90}{Cross-lingual}}     & 0                                  & 68.0$^{\pm1.1}$ & 71.1$^{\pm0.8}$ & 63.7$^{\pm0.7}$ & 70.4$^{\pm0.7}$ & 71.5$^{\pm1.1}$ & 70.5$^{\pm2.7}$ & 67.7$^{\pm1.1}$          & 71.0$^{\pm1.1}$          & 64.4$^{\pm1.3}$          & 70.4$^{\pm1.1}$          & 71.4$^{\pm0.3}$          & 70.7$^{\pm3.0}$          \\
                               &                                    & 1                                & 68.0$^{\pm0.5}$ & 71.3$^{\pm0.9}$ & 64.5$^{\pm1.3}$ & 69.6$^{\pm0.4}$ & 72.5$^{\pm1.0}$ & 71.3$^{\pm1.6}$ & 67.6$^{\pm0.3}$          & 71.7$^{\pm0.9}$          & 63.3$^{\pm0.5}$          & 69.8$^{\pm1.1}$          & 72.1$^{\pm0.3}$          & 72.3$^{\pm1.6}$          \\
                               &                                    & 2                                  & 67.2$^{\pm0.7}$ & 72.1$^{\pm0.7}$ & 63.8$^{\pm0.9}$ & 69.9$^{\pm0.6}$ & 71.7$^{\pm0.6}$ & 70.0$^{\pm1.5}$ & 67.5$^{\pm0.2}$          & 71.4$^{\pm0.1}$          & 64.4$^{\pm1.3}$          & 70.0$^{\pm1.3}$          & 72.2$^{\pm0.8}$          & 71.9$^{\pm1.5}$          \\
                               &                                    & 5                                  & 66.8$^{\pm1.2}$ & 71.1$^{\pm0.5}$ & 63.4$^{\pm1.1}$ & 69.2$^{\pm0.4}$ & 72.3$^{\pm0.7}$ & 69.5$^{\pm1.8}$ & 66.5$^{\pm0.9}$          & 71.8$^{\pm0.5}$          & 63.8$^{\pm1.3}$          & 69.0$^{\pm1.4}$          & 72.8$^{\pm0.6}$          & 70.7$^{\pm1.4}$          \\
                               &                                    & 10                                 & 68.5$^{\pm0.7}$ & 71.2$^{\pm0.8}$ & 64.4$^{\pm1.2}$ & 71.2$^{\pm1.3}$ & 72.0$^{\pm0.6}$ & 69.8$^{\pm3.2}$ & 68.2$^{\pm0.5}$          & 72.0$^{\pm0.5}$          & 64.1$^{\pm0.2}$          & 70.7$^{\pm0.9}$          & 71.8$^{\pm1.1}$          & 69.9$^{\pm1.0}$          \\
                               &                                    & 20                                 & 68.5$^{\pm0.5}$ & 71.4$^{\pm1.3}$ & 64.5$^{\pm0.8}$ & 71.9$^{\pm0.9}$ & 72.7$^{\pm1.4}$ & 68.9$^{\pm2.7}$ & 66.5$^{\pm1.6}$          & 71.7$^{\pm0.5}$          & 65.2$^{\pm1.6}$          & 71.4$^{\pm0.5}$          & 72.4$^{\pm0.7}$          & 70.1$^{\pm2.0}$          \\
                               &                                    & 100                                & 69.9$^{\pm0.6}$ & 73.9$^{\pm0.5}$ & 65.7$^{\pm0.9}$ & 71.7$^{\pm0.7}$ & 71.9$^{\pm0.9}$ & 76.0$^{\pm2.5}$ & 69.4$^{\pm0.8}$          & 73.7$^{\pm0.6}$          & 66.9$^{\pm1.2}$          & 71.1$^{\pm0.8}$          & 72.2$^{\pm0.6}$          & 73.8$^{\pm0.4}$          \\
                               \cdashlinelr{1-15}
\multirow{8}{*}{\rotatebox[origin=c]{90}{E2E-ABSA}}           & Mono                        &                                    & 73.4$^{\pm0.8}$ & *74.4$^{\pm0.6}$ & *69.9$^{\pm0.5}$ & *71.6$^{\pm1.0}$ & *72.4$^{\pm0.2}$ & *60.1$^{\pm1.7}$ & 73.5$^{\pm0.4}$          & *75.3$^{\pm0.6}$          & *69.8$^{\pm1.4}$          & *67.0$^{\pm0.4}$          & *72.2$^{\pm0.4}$          & *60.7$^{\pm1.1}$          \\  \cdashlinelr{2-15}
                               & \multirow{7}{*}{\rotatebox[origin=c]{90}{Cross-lingual}}     & 0                                  & 57.3$^{\pm1.4}$ & *59.2$^{\pm0.5}$ & *57.8$^{\pm1.2}$ & *57.1$^{\pm0.9}$ & *56.4$^{\pm2.1}$ & *44.4$^{\pm1.4}$ & \textbf{62.4}$^{\pm1.6}$ & *\textbf{69.3}$^{\pm1.0}$ & *\textbf{61.1}$^{\pm1.2}$ & *\textbf{60.8}$^{\pm0.3}$ & *\textbf{63.7}$^{\pm1.3}$ & *\textbf{48.9}$^{\pm1.4}$ \\
                               &                                    & 1                                  & 59.0$^{\pm1.1}$ & 61.1$^{\pm0.9}$ & 60.0$^{\pm1.3}$ & 59.0$^{\pm0.7}$ & 59.4$^{\pm0.8}$ & 49.2$^{\pm1.2}$ & \textbf{62.3}$^{\pm0.9}$ & \textbf{70.5}$^{\pm1.0}$ & 60.2$^{\pm0.3}$          & \textbf{61.7}$^{\pm0.9}$ & \textbf{63.4}$^{\pm1.0}$ & 51.4$^{\pm3.0}$          \\
                               &                                    & 2                                  & 58.9$^{\pm0.6}$ & 62.8$^{\pm2.7}$ & 59.3$^{\pm1.1}$ & 61.0$^{\pm0.4}$ & 59.6$^{\pm0.6}$ & 51.2$^{\pm1.9}$ & \textbf{63.0}$^{\pm0.5}$ & \textbf{71.3}$^{\pm0.7}$ & 60.4$^{\pm1.2}$          & 62.1$^{\pm1.1}$          & \textbf{63.7}$^{\pm0.9}$ & 52.8$^{\pm1.8}$          \\
                               &                                    & 5                                  & 59.9$^{\pm1.1}$ & 63.2$^{\pm2.4}$ & 61.3$^{\pm1.1}$ & 60.5$^{\pm1.0}$ & 61.2$^{\pm0.7}$ & 54.8$^{\pm3.0}$ & \textbf{63.2}$^{\pm0.3}$ & \textbf{71.0}$^{\pm1.6}$ & 61.4$^{\pm0.6}$          & 61.1$^{\pm2.0}$          & \textbf{65.8}$^{\pm0.8}$ & 53.2$^{\pm0.9}$          \\
                               &                                    & 10                                 & 60.7$^{\pm1.0}$ & 70.7$^{\pm1.3}$ & 60.9$^{\pm0.6}$ & 61.9$^{\pm1.3}$ & 60.0$^{\pm1.3}$ & 53.2$^{\pm1.7}$ & \textbf{63.6}$^{\pm0.7}$ & 71.6$^{\pm0.9}$          & 61.4$^{\pm0.7}$          & 62.7$^{\pm1.5}$          & \textbf{65.8}$^{\pm1.2}$ & 54.0$^{\pm1.6}$          \\
                               &                                    & 20                                 & 62.1$^{\pm1.2}$ & 71.2$^{\pm1.7}$ & 61.1$^{\pm0.7}$ & 62.1$^{\pm1.5}$ & 64.6$^{\pm1.7}$ & 52.5$^{\pm1.9}$ & 64.0$^{\pm1.6}$          & 72.1$^{\pm1.0}$          & 62.2$^{\pm1.0}$          & 63.2$^{\pm1.3}$          & 66.6$^{\pm1.0}$          & 56.2$^{\pm2.7}$          \\
                               &                                    & 100                                & 66.6$^{\pm0.4}$ & 72.2$^{\pm0.5}$ & 62.7$^{\pm1.0}$ & 65.8$^{\pm1.6}$ & 66.0$^{\pm1.6}$ & 57.7$^{\pm1.8}$ & \textbf{68.0}$^{\pm0.6}$ & \textbf{73.1}$^{\pm0.3}$ & 62.3$^{\pm0.4}$          & 64.9$^{\pm0.5}$          & 67.6$^{\pm0.8}$          & 58.4$^{\pm1.6}$          \\ \cdashlinelr{1-15}
\multirow{8}{*}{\rotatebox[origin=c]{90}{ACTE}}          & Mono                        &                                    & 73.5$^{\pm0.8}$ & *70.4$^{\pm0.7}$ & *63.7$^{\pm0.8}$ & *68.8$^{\pm0.5}$ & *73.2$^{\pm0.5}$ & *59.1$^{\pm0.5}$ & 73.6$^{\pm0.5}$          & *69.9$^{\pm0.4}$          & *64.9$^{\pm0.5}$          & *62.9$^{\pm0.5}$          & *72.8$^{\pm1.0}$          & *60.4$^{\pm2.1}$          \\  \cdashlinelr{2-15}
                               & \multirow{7}{*}{\rotatebox[origin=c]{90}{Cross-lingual}}     & 0                                  & 54.3$^{\pm1.6}$ & *52.5$^{\pm1.0}$ & *55.8$^{\pm0.7}$ & *52.3$^{\pm1.3}$ & *55.0$^{\pm2.7}$ & *41.4$^{\pm1.4}$ & \textbf{58.7}$^{\pm1.0}$ & *\textbf{62.8}$^{\pm1.4}$ & *\textbf{57.5}$^{\pm0.3}$ & *\textbf{54.1}$^{\pm0.2}$ & *\textbf{60.4}$^{\pm0.9}$ & *\textbf{49.0}$^{\pm0.9}$ \\
                               &                                    & 1                                  & 55.2$^{\pm0.9}$ & 58.5$^{\pm0.8}$ & 56.2$^{\pm0.3}$ & 52.7$^{\pm1.1}$ & 57.1$^{\pm2.1}$ & 45.0$^{\pm2.1}$ & \textbf{58.8}$^{\pm0.8}$ & \textbf{62.8}$^{\pm0.2}$ & 55.8$^{\pm0.5}$          & 53.9$^{\pm0.8}$          & \textbf{61.5}$^{\pm0.5}$ & 46.9$^{\pm1.1}$          \\
                               &                                    & 2                                  & 55.8$^{\pm1.2}$ & 57.7$^{\pm1.8}$ & 55.0$^{\pm0.7}$ & 52.1$^{\pm0.6}$ & 57.5$^{\pm1.1}$ & 46.4$^{\pm0.7}$ & \textbf{58.9}$^{\pm1.0}$ & \textbf{62.8}$^{\pm0.5}$ & \textbf{57.1}$^{\pm0.9}$ & 53.4$^{\pm0.9}$          & \textbf{62.0}$^{\pm0.4}$ & 47.1$^{\pm0.8}$          \\
                               &                                    & 5                                  & 57.0$^{\pm1.0}$ & 58.6$^{\pm0.4}$ & 56.3$^{\pm0.4}$ & 53.9$^{\pm0.9}$ & 58.1$^{\pm1.0}$ & 48.0$^{\pm3.4}$ & \textbf{59.7}$^{\pm0.8}$ & \textbf{63.2}$^{\pm1.0}$ & \textbf{58.0}$^{\pm0.7}$ & 53.4$^{\pm1.6}$          & \textbf{62.8}$^{\pm0.6}$ & 48.9$^{\pm1.9}$          \\
                               &                                    & 10                                 & 56.8$^{\pm1.3}$ & 63.3$^{\pm0.3}$ & 57.4$^{\pm1.6}$ & 56.1$^{\pm1.2}$ & 59.3$^{\pm1.3}$ & 49.7$^{\pm0.7}$ & \textbf{60.3}$^{\pm1.3}$ & 63.9$^{\pm0.9}$          & 57.0$^{\pm0.8}$          & 55.7$^{\pm1.7}$          & \textbf{63.2}$^{\pm0.8}$ & 50.7$^{\pm1.3}$          \\
                               &                                    & 20                                 & 57.4$^{\pm1.6}$ & 64.6$^{\pm0.5}$ & 57.8$^{\pm0.8}$ & 55.2$^{\pm1.2}$ & 63.5$^{\pm0.8}$ & 49.8$^{\pm1.2}$ & \textbf{61.0}$^{\pm1.1}$ & 64.5$^{\pm1.0}$          & 57.3$^{\pm0.7}$          & 55.5$^{\pm0.5}$          & 64.2$^{\pm1.2}$          & 51.1$^{\pm1.7}$          \\
                               &                                    & 100                                & 63.6$^{\pm0.7}$ & 65.9$^{\pm0.6}$ & 59.8$^{\pm0.4}$ & 60.1$^{\pm1.0}$ & 66.0$^{\pm0.7}$ & 53.6$^{\pm1.2}$ & 64.4$^{\pm1.1}$          & 66.4$^{\pm1.0}$          & 59.7$^{\pm0.9}$          & 58.5$^{\pm0.5}$          & 66.2$^{\pm0.8}$          & 55.2$^{\pm1.6}$          \\ \cdashlinelr{1-15}
\multirow{8}{*}{\rotatebox[origin=c]{90}{TASD}}          & Mono                       &                                    & 66.9$^{\pm0.3}$ & *65.8$^{\pm0.4}$ & *59.0$^{\pm0.6}$ & *62.9$^{\pm1.4}$ & *67.0$^{\pm0.9}$ & *54.1$^{\pm3.0}$ & 67.1$^{\pm1.3}$          & *66.2$^{\pm0.5}$          & *58.9$^{\pm1.1}$          & *57.6$^{\pm0.5}$          & *66.4$^{\pm0.4}$          & *53.9$^{\pm1.5}$          \\ \cdashlinelr{2-15}
                               & \multirow{7}{*}{\rotatebox[origin=c]{90}{Cross-lingual}}     & 0                                  & 50.2$^{\pm0.9}$ & *48.3$^{\pm0.5}$ & *50.4$^{\pm1.4}$ & *47.7$^{\pm1.1}$ & *48.6$^{\pm2.0}$ & *39.1$^{\pm3.6}$ & \textbf{53.3}$^{\pm1.5}$ & *\textbf{57.6}$^{\pm0.6}$ & *50.4$^{\pm0.8}$          & *\textbf{50.4}$^{\pm1.3}$ & *\textbf{54.9}$^{\pm2.0}$ & *\textbf{43.8}$^{\pm0.8}$ \\
                               &                                    & 1                                  & 49.9$^{\pm0.4}$ & 52.5$^{\pm1.2}$ & 50.0$^{\pm1.2}$ & 48.1$^{\pm0.8}$ & 52.3$^{\pm1.6}$ & 42.0$^{\pm1.8}$ & \textbf{53.6}$^{\pm0.9}$ & \textbf{58.0}$^{\pm0.5}$ & 50.4$^{\pm1.0}$          & 49.8$^{\pm0.9}$          & \textbf{55.5}$^{\pm0.5}$ & 43.9$^{\pm1.6}$          \\
                               &                                    & 2                                  & 50.7$^{\pm1.5}$ & 51.3$^{\pm1.5}$ & 49.6$^{\pm1.2}$ & 49.1$^{\pm0.6}$ & 53.1$^{\pm1.5}$ & 43.7$^{\pm1.9}$ & \textbf{53.9}$^{\pm0.9}$ & \textbf{58.0}$^{\pm0.6}$ & 50.9$^{\pm1.4}$          & \textbf{50.2}$^{\pm0.3}$ & \textbf{55.5}$^{\pm0.7}$ & 44.1$^{\pm1.7}$          \\
                               &                                    & 5                                  & 52.4$^{\pm1.7}$ & 52.4$^{\pm2.1}$ & 50.2$^{\pm1.2}$ & 49.7$^{\pm1.3}$ & 52.3$^{\pm1.4}$ & 45.1$^{\pm1.4}$ & 53.2$^{\pm1.1}$          & \textbf{58.4}$^{\pm0.6}$ & 51.2$^{\pm1.6}$          & 49.0$^{\pm1.3}$          & \textbf{57.5}$^{\pm0.4}$ & 46.5$^{\pm1.0}$          \\
                               &                                    & 10                                 & 52.5$^{\pm1.3}$ & 58.2$^{\pm1.0}$ & 49.8$^{\pm0.9}$ & 51.2$^{\pm1.5}$ & 53.6$^{\pm1.5}$ & 45.3$^{\pm2.7}$ & 54.1$^{\pm0.6}$          & 58.8$^{\pm0.6}$          & 51.0$^{\pm0.8}$          & 50.4$^{\pm1.2}$          & \textbf{57.5}$^{\pm1.1}$ & 46.3$^{\pm1.5}$          \\
                               &                                    & 20                                 & 51.6$^{\pm0.6}$ & 58.9$^{\pm0.8}$ & 50.6$^{\pm1.3}$ & 52.5$^{\pm0.8}$ & 57.8$^{\pm1.0}$ & 44.8$^{\pm2.1}$ & \textbf{54.1}$^{\pm1.3}$ & 60.2$^{\pm0.7}$          & 52.5$^{\pm1.4}$          & 52.3$^{\pm0.8}$          & 58.0$^{\pm1.3}$          & 47.4$^{\pm2.2}$          \\
                               &                                    & 100                                & 57.5$^{\pm0.4}$ & 61.5$^{\pm0.8}$ & 52.4$^{\pm1.1}$ & 54.1$^{\pm0.7}$ & 58.9$^{\pm0.9}$ & 49.7$^{\pm1.4}$ & 57.1$^{\pm0.7}$          & 61.7$^{\pm0.7}$          & 52.7$^{\pm1.6}$          & 54.3$^{\pm0.8}$          & 58.8$^{\pm1.4}$          & 50.1$^{\pm0.6}$          \\ \bottomrule
\end{tabular}
\end{adjustbox}
\label{tab:res_mt5}
\end{table}

Overall, we observe a clear trend of performance improvement with an increasing number of few-shot examples. This improvement is especially noticeable when constrained decoding is not used. For instance, in the TASD task with Spanish as the target language, adding a single few-shot example improves performance by approximately 4\% over the zero-shot setting without constrained decoding. Increasing the number of examples to ten provides an additional 6\% gain -- a 10\% improvement over the zero-shot baseline. Interestingly, as the number of few-shot examples increases, the relative benefit of constrained decoding decreases. With ten few-shot examples, constrained decoding usually provides no significant advantage. An exception is Czech, where constrained decoding consistently improves results, even with 100 few-shot examples -- possibly due to specific morphological or syntactic characteristics of the language.

The ACSA task, which had not previously been evaluated with constrained decoding, shows no consistent improvement from its use. This aligns with expectations, as constrained decoding primarily addresses challenges related to aspect term prediction~\cite{icaart25}, which is not a component of ACSA. 

Across tasks, the most substantial gains compared to zero-shot performance generally occur when moving to ten few-shot examples. Improvements with fewer examples (e.g. one or five) tend to be modest or not statistically significant. Increasing the count from ten to twenty examples typically yields to only a small or no further gains. A more substantial leap in performance is usually seen only when scaling up to 100 examples, suggesting a non-linear benefit from additional supervision. With 100 few-shot examples, performance is typically between 2–-10\% lower than monolingual results, depending on the task, while with ten few-shot examples, performance is usually about 3\% lower than with 100 examples.

\begin{table}[ht!]
\centering
\caption{Cross-lingual micro F1-scores on four tasks in six target languages using mBART, compared to monolingual (mono) performance, with varying few-shot examples (FS). \textbf{Bold} marks better constrained decoding (CD) results with non-overlapping 95\% confidence intervals. Asterisked (*) results are from~\cite{icaart25}.}
\begin{adjustbox}{width=\linewidth}
\begin{tabular}{@{}ccr@{\hspace{6pt}}rrrrrrrrrrrr@{}}
\toprule
\multirow{2}{*}{\textbf{Task}} & \multirow{2}{*}{\textbf{Setup}} & \multicolumn{1}{l}{\multirow{2}{*}{\textbf{FS}}} & \multicolumn{6}{c}{\textbf{Without constrained decoding}}                                                                                           & \multicolumn{6}{c}{\textbf{With constrained decoding}}                                                                                                      \\ \cmidrule(lr){4-9}  \cmidrule(lr){10-15} 
                               &                                    & \multicolumn{1}{l}{}                                   & \multicolumn{1}{c}{Cs} & \multicolumn{1}{c}{Es} & \multicolumn{1}{c}{Fr} & \multicolumn{1}{c}{Nl} & \multicolumn{1}{c}{Ru} & \multicolumn{1}{c}{Tr} & \multicolumn{1}{c}{Cs} & \multicolumn{1}{c}{Es}   & \multicolumn{1}{c}{Fr} & \multicolumn{1}{c}{Nl}   & \multicolumn{1}{c}{Ru}   & \multicolumn{1}{c}{Tr}   \\ \midrule
\multirow{8}{*}{\rotatebox[origin=c]{90}{ACSA}}          & Mono                        & \multicolumn{1}{l}{}                                   & 72.6$^{\pm1.3}$        & 73.2$^{\pm1.4}$        & 65.0$^{\pm0.9}$        & 70.2$^{\pm1.9}$        & 73.3$^{\pm1.2}$        & 66.1$^{\pm4.0}$        & 71.5$^{\pm0.8}$        & 73.3$^{\pm0.8}$          & 63.7$^{\pm1.0}$        & 68.3$^{\pm2.9}$          & 72.8$^{\pm1.0}$          & 64.4$^{\pm5.1}$          \\ \cdashlinelr{2-15}
                               & \multirow{7}{*}{\rotatebox[origin=c]{90}{Cross-lingual}}     & 0                                                      & 55.2$^{\pm1.4}$        & 61.7$^{\pm2.5}$        & 53.7$^{\pm2.4}$        & 58.4$^{\pm2.1}$        & 65.6$^{\pm1.9}$        & 50.7$^{\pm2.6}$        & 53.8$^{\pm4.0}$        & 63.0$^{\pm2.3}$          & 53.0$^{\pm1.8}$        & 59.4$^{\pm2.0}$          & 66.4$^{\pm1.7}$          & 49.3$^{\pm2.7}$          \\
                               &                                    & 1                                                      & 56.6$^{\pm1.2}$        & 64.1$^{\pm1.0}$        & 54.9$^{\pm1.0}$        & 60.1$^{\pm0.9}$        & 68.2$^{\pm0.9}$        & 54.4$^{\pm1.1}$        & 57.4$^{\pm1.1}$        & 64.2$^{\pm0.5}$          & 54.0$^{\pm0.8}$        & 61.1$^{\pm1.8}$          & 68.5$^{\pm1.1}$          & 54.8$^{\pm0.7}$          \\
                               &                                    & 2                                                      & 56.7$^{\pm1.4}$        & 64.4$^{\pm1.1}$        & 55.7$^{\pm1.1}$        & 60.9$^{\pm0.8}$        & 67.2$^{\pm1.3}$        & 52.1$^{\pm1.7}$        & 56.3$^{\pm1.6}$        & 65.2$^{\pm1.3}$          & 55.0$^{\pm1.5}$        & 60.8$^{\pm0.9}$          & 68.1$^{\pm0.8}$          & 53.2$^{\pm2.0}$          \\
                               &                                    & 5                                                      & 56.4$^{\pm0.9}$        & 63.9$^{\pm1.8}$        & 56.2$^{\pm1.2}$        & 61.3$^{\pm1.3}$        & 67.1$^{\pm0.6}$        & 51.6$^{\pm4.8}$        & 56.2$^{\pm0.7}$        & 65.0$^{\pm1.4}$          & 55.9$^{\pm2.0}$        & 61.5$^{\pm1.1}$          & 67.6$^{\pm0.7}$          & 53.7$^{\pm3.5}$          \\
                               &                                    & 10                                                     & 57.9$^{\pm2.5}$        & 62.5$^{\pm3.1}$        & 55.7$^{\pm1.3}$        & 58.4$^{\pm2.2}$        & 66.5$^{\pm1.8}$        & 54.4$^{\pm1.9}$        & 55.5$^{\pm1.5}$        & 62.8$^{\pm2.0}$          & 54.0$^{\pm3.3}$        & 58.3$^{\pm3.9}$          & 67.6$^{\pm0.5}$          & 53.4$^{\pm5.8}$          \\
                               &                                    & 20                                                     & 57.0$^{\pm3.4}$        & 64.0$^{\pm2.3}$        & 57.2$^{\pm2.4}$        & 62.8$^{\pm0.7}$        & 65.9$^{\pm1.5}$        & 55.5$^{\pm3.2}$        & 56.0$^{\pm4.1}$        & 61.5$^{\pm1.8}$          & 56.9$^{\pm2.1}$        & 60.3$^{\pm3.7}$          & 66.0$^{\pm0.7}$          & 55.9$^{\pm4.3}$          \\
                               &                                    & 100                                                    & 63.5$^{\pm1.5}$        & 67.6$^{\pm2.1}$        & 59.4$^{\pm2.2}$        & 64.5$^{\pm1.9}$        & 67.1$^{\pm3.1}$        & 61.9$^{\pm4.6}$        & 62.4$^{\pm2.0}$        & 66.5$^{\pm0.9}$          & 59.3$^{\pm1.5}$        & 65.0$^{\pm1.6}$          & 65.5$^{\pm2.3}$          & 62.6$^{\pm4.7}$          \\ \cdashlinelr{1-15}
\multirow{8}{*}{\rotatebox[origin=c]{90}{E2E-ABSA}}           & Mono                        & \multicolumn{1}{l}{}                                   & 69.1$^{\pm0.3}$        & *73.0$^{\pm0.5}$        & *66.4$^{\pm1.1}$        & *68.9$^{\pm1.2}$        & *68.7$^{\pm1.6}$        & *56.0$^{\pm2.7}$        & 68.8$^{\pm0.8}$        & *71.9$^{\pm1.3}$          & *64.0$^{\pm1.7}$        & *61.6$^{\pm1.0}$          & *66.2$^{\pm1.1}$          & *54.4$^{\pm2.3}$          \\ \cdashlinelr{2-15}
                               & \multirow{7}{*}{\rotatebox[origin=c]{90}{Cross-lingual}}     & 0                                                      & 51.5$^{\pm3.3}$        & *61.1$^{\pm2.6}$        & *49.4$^{\pm3.8}$        & *51.6$^{\pm2.7}$        & *57.1$^{\pm1.4}$        & *31.6$^{\pm3.9}$        & 48.8$^{\pm2.7}$        & *61.7$^{\pm2.7}$          & *49.2$^{\pm4.1}$        & *50.1$^{\pm3.5}$          & *57.8$^{\pm1.8}$          & *30.3$^{\pm3.0}$          \\
                               &                                    & 1                                                      & 49.2$^{\pm1.2}$        & 56.6$^{\pm0.6}$        & 49.9$^{\pm0.5}$        & 47.3$^{\pm0.7}$        & 55.8$^{\pm1.2}$        & 31.8$^{\pm1.5}$        & 49.7$^{\pm0.8}$        & 57.1$^{\pm1.6}$          & 48.9$^{\pm1.1}$        & 46.1$^{\pm0.7}$          & 56.9$^{\pm1.2}$          & 32.6$^{\pm2.4}$          \\
                               &                                    & 2                                                      & 50.1$^{\pm2.0}$        & 55.7$^{\pm1.1}$        & 50.1$^{\pm1.6}$        & 47.7$^{\pm1.1}$        & 57.5$^{\pm1.6}$        & 32.6$^{\pm2.6}$        & 49.4$^{\pm1.6}$        & 56.5$^{\pm0.6}$          & 48.6$^{\pm0.5}$        & 47.9$^{\pm1.8}$          & 56.7$^{\pm1.2}$          & 32.7$^{\pm1.6}$          \\
                               &                                    & 5                                                      & 51.8$^{\pm0.7}$        & 57.0$^{\pm1.1}$        & 51.3$^{\pm1.0}$        & 49.4$^{\pm0.6}$        & 57.5$^{\pm1.2}$        & 30.7$^{\pm0.8}$        & 50.6$^{\pm1.4}$        & 56.4$^{\pm0.8}$          & 50.2$^{\pm1.5}$        & 47.9$^{\pm1.4}$          & 56.2$^{\pm0.4}$          & 31.7$^{\pm2.3}$          \\
                               &                                    & 10                                                     & 51.0$^{\pm3.4}$        & 64.1$^{\pm2.2}$        & 53.4$^{\pm2.0}$        & 53.4$^{\pm2.9}$        & 58.8$^{\pm1.7}$        & 37.7$^{\pm3.2}$        & 52.0$^{\pm3.4}$        & 64.8$^{\pm2.6}$          & 51.6$^{\pm1.1}$        & 51.4$^{\pm2.9}$          & 57.9$^{\pm1.8}$          & 36.3$^{\pm2.0}$          \\
                               &                                    & 20                                                     & 53.6$^{\pm2.9}$        & 63.2$^{\pm2.8}$        & 54.3$^{\pm1.8}$        & 53.4$^{\pm2.0}$        & 59.5$^{\pm1.6}$        & 40.9$^{\pm3.9}$        & 52.8$^{\pm3.4}$        & 65.5$^{\pm1.3}$          & 51.7$^{\pm1.1}$        & 53.6$^{\pm1.2}$          & 58.9$^{\pm2.4}$          & 36.4$^{\pm1.9}$          \\
                               &                                    & 100                                                    & 60.6$^{\pm1.3}$        & 67.6$^{\pm1.4}$        & 57.2$^{\pm1.2}$        & 57.9$^{\pm1.7}$        & 62.3$^{\pm1.9}$        & 47.2$^{\pm3.1}$        & 59.7$^{\pm1.7}$        & 67.5$^{\pm1.1}$          & 55.3$^{\pm0.8}$        & 56.1$^{\pm1.6}$          & 60.5$^{\pm0.8}$          & 46.8$^{\pm3.7}$          \\ \cdashlinelr{1-15}
\multirow{8}{*}{\rotatebox[origin=c]{90}{ACTE}}          & Mono                        & \multicolumn{1}{l}{}                                   & 70.1$^{\pm0.8}$        & *66.4$^{\pm1.6}$        & *61.1$^{\pm1.6}$        & *64.1$^{\pm1.2}$        & *70.9$^{\pm0.6}$        & *56.8$^{\pm2.2}$        & 68.4$^{\pm0.9}$        & *66.8$^{\pm1.5}$          & *58.2$^{\pm1.2}$        & *58.0$^{\pm1.2}$          & *67.4$^{\pm0.3}$          & *55.3$^{\pm1.5}$          \\ \cdashlinelr{2-15}
                               & \multirow{7}{*}{\rotatebox[origin=c]{90}{Cross-lingual}}     & 0                                                      & 48.6$^{\pm2.7}$        & *52.5$^{\pm1.4}$        & *49.3$^{\pm1.5}$        & *44.5$^{\pm1.4}$        & *53.8$^{\pm1.5}$        & *31.1$^{\pm2.1}$        & 45.8$^{\pm4.5}$        & *\textbf{54.8}$^{\pm0.4}$ & *49.2$^{\pm0.6}$        & *\textbf{46.9}$^{\pm0.9}$ & *\textbf{55.9}$^{\pm0.2}$ & *\textbf{34.7}$^{\pm1.1}$ \\
                               &                                    & 1                                                      & 52.7$^{\pm1.0}$        & 63.0$^{\pm1.0}$        & 52.3$^{\pm2.1}$        & 52.2$^{\pm0.9}$        & 58.0$^{\pm0.7}$        & 33.6$^{\pm2.2}$        & 52.1$^{\pm1.8}$        & 64.5$^{\pm1.1}$          & 51.8$^{\pm1.3}$        & 52.6$^{\pm0.9}$          & 58.8$^{\pm1.7}$          & 34.5$^{\pm1.6}$          \\
                               &                                    & 2                                                      & 52.7$^{\pm1.2}$        & 63.3$^{\pm0.7}$        & 51.6$^{\pm0.9}$        & 52.1$^{\pm2.2}$        & 59.2$^{\pm0.8}$        & 35.2$^{\pm1.0}$        & 51.8$^{\pm2.0}$        & \textbf{64.7}$^{\pm0.5}$ & 50.8$^{\pm1.0}$        & 52.1$^{\pm1.4}$          & 59.0$^{\pm0.8}$          & 34.7$^{\pm1.2}$          \\
                               &                                    & 5                                                      & 50.9$^{\pm0.9}$        & 63.6$^{\pm0.6}$        & 54.1$^{\pm1.0}$        & 52.5$^{\pm0.5}$        & 59.6$^{\pm0.7}$        & 36.4$^{\pm1.0}$        & 51.8$^{\pm2.1}$        & 64.9$^{\pm0.7}$          & 53.4$^{\pm1.5}$        & 51.9$^{\pm0.5}$          & 59.1$^{\pm0.5}$          & 35.5$^{\pm0.9}$          \\
                               &                                    & 10                                                     & 52.5$^{\pm0.9}$        & 57.3$^{\pm1.1}$        & 50.6$^{\pm1.9}$        & 47.4$^{\pm2.2}$        & 56.0$^{\pm2.3}$        & 35.1$^{\pm2.9}$        & 47.5$^{\pm0.8}$        & 56.4$^{\pm1.5}$          & 48.5$^{\pm0.5}$        & 45.9$^{\pm1.3}$          & 55.7$^{\pm1.9}$          & 33.6$^{\pm5.1}$          \\
                               &                                    & 20                                                     & 50.8$^{\pm1.0}$        & 59.4$^{\pm0.6}$        & 50.9$^{\pm1.5}$        & 49.8$^{\pm2.1}$        & 59.7$^{\pm0.9}$        & 41.1$^{\pm2.9}$        & 51.6$^{\pm2.8}$        & 57.8$^{\pm1.6}$          & 47.8$^{\pm1.6}$        & 48.8$^{\pm2.8}$          & 59.1$^{\pm1.6}$          & 39.6$^{\pm4.4}$          \\
                               &                                    & 100                                                    & 59.2$^{\pm1.6}$        & 60.4$^{\pm2.5}$        & 53.3$^{\pm2.2}$        & 55.0$^{\pm2.4}$        & 62.1$^{\pm1.0}$        & 46.6$^{\pm4.9}$        & 57.1$^{\pm2.2}$        & 61.0$^{\pm1.8}$          & 51.1$^{\pm1.1}$        & 52.5$^{\pm2.1}$          & 60.3$^{\pm1.9}$          & 44.8$^{\pm3.9}$          \\ \cdashlinelr{1-15}
\multirow{8}{*}{\rotatebox[origin=c]{90}{TASD}}          & Mono                        & \multicolumn{1}{l}{}                                   & 62.6$^{\pm0.7}$        & *62.9$^{\pm1.2}$        & *54.8$^{\pm0.9}$        & *57.6$^{\pm0.9}$        & *62.6$^{\pm0.7}$        & *49.3$^{\pm3.1}$        & 61.9$^{\pm1.6}$        & *61.5$^{\pm1.4}$          & *52.4$^{\pm0.6}$        & *52.1$^{\pm1.0}$          & *60.1$^{\pm1.9}$          & *47.6$^{\pm2.7}$          \\ \cdashlinelr{2-15}
                               & \multirow{7}{*}{\rotatebox[origin=c]{90}{Cross-lingual}}     & 0                                                      & 40.4$^{\pm3.0}$        & *47.6$^{\pm1.9}$        & *39.6$^{\pm0.8}$        & *39.1$^{\pm0.9}$        & *48.5$^{\pm1.1}$        & *23.5$^{\pm2.6}$        & 39.3$^{\pm1.0}$        & *\textbf{51.1}$^{\pm1.2}$ & *39.9$^{\pm0.6}$        & *38.9$^{\pm0.9}$          & *\textbf{50.5}$^{\pm0.7}$ & *\textbf{27.3}$^{\pm1.1}$ \\
                               &                                    & 1                                                      & 42.8$^{\pm1.2}$        & 50.4$^{\pm1.8}$        & 41.3$^{\pm1.2}$        & 40.9$^{\pm0.8}$        & 51.2$^{\pm0.4}$        & 26.8$^{\pm1.2}$        & 41.1$^{\pm1.0}$        & 51.5$^{\pm0.9}$          & 40.5$^{\pm0.9}$        & 39.2$^{\pm0.7}$          & 50.8$^{\pm0.9}$          & 27.2$^{\pm1.4}$          \\
                               &                                    & 2                                                      & 42.4$^{\pm1.2}$        & 50.5$^{\pm1.0}$        & 42.3$^{\pm0.9}$        & 40.6$^{\pm1.1}$        & 50.8$^{\pm0.7}$        & 28.9$^{\pm3.3}$        & 42.0$^{\pm0.9}$        & 51.3$^{\pm1.3}$          & 39.7$^{\pm1.1}$        & 40.1$^{\pm2.1}$          & 50.6$^{\pm1.2}$          & 29.1$^{\pm2.7}$          \\
                               &                                    & 5                                                      & 42.8$^{\pm1.5}$        & 52.3$^{\pm1.5}$        & 42.1$^{\pm1.2}$        & 41.6$^{\pm1.1}$        & 50.0$^{\pm0.2}$        & 27.3$^{\pm3.2}$        & 42.2$^{\pm1.9}$        & 51.3$^{\pm1.0}$          & 41.0$^{\pm0.8}$        & 41.8$^{\pm1.1}$          & \textbf{51.5}$^{\pm0.9}$ & 26.8$^{\pm1.4}$          \\
                               &                                    & 10                                                     & 43.3$^{\pm2.3}$        & 51.8$^{\pm1.0}$        & 40.9$^{\pm3.8}$        & 42.2$^{\pm3.1}$        & 49.9$^{\pm1.0}$        & 29.8$^{\pm4.4}$        & 42.4$^{\pm2.2}$        & 51.7$^{\pm3.3}$          & 40.1$^{\pm1.7}$        & 39.7$^{\pm3.5}$          & 50.0$^{\pm0.7}$          & 28.3$^{\pm1.8}$          \\
                               &                                    & 20                                                     & 43.5$^{\pm1.6}$        & 53.8$^{\pm2.2}$        & 42.0$^{\pm0.6}$        & 42.1$^{\pm2.4}$        & 51.6$^{\pm2.2}$        & 32.2$^{\pm0.6}$        & 42.0$^{\pm3.1}$        & 51.4$^{\pm1.8}$          & 40.3$^{\pm2.5}$        & 41.3$^{\pm2.1}$          & 51.5$^{\pm2.0}$          & 29.9$^{\pm2.5}$          \\
                               &                                    & 100                                                    & 51.3$^{\pm0.9}$        & 56.1$^{\pm2.7}$        & 44.9$^{\pm1.2}$        & 48.2$^{\pm2.4}$        & 54.3$^{\pm0.7}$        & 40.8$^{\pm2.9}$        & 49.7$^{\pm1.0}$        & 55.8$^{\pm0.3}$          & 42.4$^{\pm1.8}$        & 46.8$^{\pm1.9}$          & 52.0$^{\pm1.2}$          & 39.8$^{\pm1.8}$          \\ \bottomrule
\end{tabular}
\end{adjustbox}
\label{tab:res_mbart}
\end{table}

Table~\ref{tab:res_mbart} shows analogous results using the mBART model. While the overall patterns mirror those observed with mT5, a notable difference emerges: constrained decoding has significantly less impact on mBART. It generally leads to improvements only in a subset of languages under zero-shot settings. This suggests that mBART may be inherently more robust to output structure violations or benefits less from structural constraints. However, the overall results are better with mT5 than with mBART.

In summary, even a small number of high-quality few-shot examples -- particularly ten -- can yield substantial gains over zero-shot performance, often surpassing zero-shot constrained decoding results. Given that collecting ten labelled instances per target language is a manageable effort, few-shot learning presents a highly practical and efficient approach for cross-lingual ABSA. Moreover, the diminishing returns beyond ten examples -- particularly when weighed against the increased time and cost of data labelling -- underscore the efficiency of small-scale supervision and offer promising implications for low-resource or domain-specific adaptation scenarios.

\subsection{TASD with Different Few-Shot Examples}

\begin{figure}[ht!]
    \centering
    \begin{subfigure}[b]{0.32\textwidth}
        \centering
        \begin{adjustbox}{width=\textwidth}
            \begin{tikzpicture}
                \begin{axis}[
                    xlabel={{Number of target language examples}},
                    ylabel={\small{F1-score [\%]}},
                    xmin=0, xmax=ALL,
                    symbolic x coords={0, 1, 2, 5, 10, 20, 100, 200, 500, {1,000}, ALL},
                    xtick=data,
                    ymin=37, ymax=69,
                    ytick={40, 45, 50, 55, 60, 65},
                    legend pos=south east,
                    ymajorgrids=true,
                    xmajorgrids=true,
                    grid style={black, dotted},
                    ylabel near ticks,
                    axis line style={black},
                    tick style={black},
                    label style={black},
                    ticklabel style={black},
                    xticklabel style={rotate=45},
                    legend style={font=\small, text=black, legend columns=2},
                    legend to name={mylegend},
                    ]
            
                    \addplot[
                        dashed,
                        color=orange,
                        mark=None, 
                        ultra thick,
                    ] 
                    coordinates {
                        (0,67.07545638)
                        (1,67.07545638)
                        (2,67.07545638)
                        (5,67.07545638)
                        (10,67.07545638)
                        (20,67.07545638)
                        (100,67.07545638)
                        (200,67.07545638)
                        (500,67.07545638)
                        ({1,000},67.07545638)
                        (ALL,67.07545638)
                    };
            
                    \addplot[
                        dotted,
                        color=green,
                        mark=None, 
                        ultra thick,
                    ] 
                    coordinates {
                        (0,66.93234921)
                        (1,66.93234921)
                        (2,66.93234921)
                        (5,66.93234921)
                        (10,66.93234921)
                        (20,66.93234921)
                        (100,66.93234921)
                        (200,66.93234921)
                        (500,66.93234921)
                        ({1,000},66.93234921)
                        (ALL,66.93234921)
                    };
            
                    \addplot[
                        color=red,
                        mark=*,
                        ultra thick,
                    ]
                    coordinates {
                        (0,53.31776023)
                        (1,53.61512064933770)
                        (2,53.88654589653010)
                        (5,53.19945693016050)
                        (10,54.06018019)
                        (20,54.11482334)
                        (100,57.10698605)
                        (200,59.95276213)
                        (500,63.14447522)
                        ({1,000},64.56431031)
                        (ALL,65.99839687)
                    };
            
                    \addplot[
                        color=blue,
                        mark=triangle*,
                        ultra thick,
                    ]
                    coordinates {
                        (0,50.2046752)
                        (1,49.89663541316980)
                        (2,50.69463968276970)
                        (5,52.35179662704470)
                        (10,52.47662663)
                        (20,51.59751058)
                        (100,57.46541739)
                        (200,59.56181169)
                        (500,62.91004062)
                        ({1,000},64.93930101)
                        (ALL,66.85097694)
                    };

                    \legend{With CD (monolingual), Without CD (monolingual), With CD (cross-lingual), Without CD (cross-lingual)}
                    
                \end{axis}
            \end{tikzpicture}
        \end{adjustbox}
        \caption{Czech}
    \end{subfigure}
    \hfill
    \begin{subfigure}[b]{0.32\textwidth}
        \centering
        \begin{adjustbox}{width=\textwidth}
            \begin{tikzpicture}
                \begin{axis}[
                    xlabel={{Number of target language examples}},
                    ylabel={\small{F1-score [\%]}},
                    xmin=0, xmax=ALL,
                    symbolic x coords={0, 1, 2, 5, 10, 20, 100, 200, 500, {1,000}, ALL},
                    xtick=data,
                    ymin=43, ymax=69,
                    ytick={40, 45, 50, 55, 60, 65},
                    legend pos=south east,
                    ymajorgrids=true,
                    xmajorgrids=true,
                    grid style={black, dotted},
                    ylabel near ticks,
                    axis line style={black},
                    tick style={black},
                    label style={black},
                    ticklabel style={black},
                    xticklabel style={rotate=45},
                    legend style={font=\footnotesize, text=black},
                    ]
            
                    \addplot[
                        dashed,
                        color=orange,
                        mark=None, 
                        ultra thick,
                    ] 
                    coordinates {
                        (0,66.23306394)
                        (1,66.23306394)
                        (2,66.23306394)
                        (5,66.23306394)
                        (10,66.23306394)
                        (20,66.23306394)
                        (100,66.23306394)
                        (200,66.23306394)
                        (500,66.23306394)
                        ({1,000},66.23306394)
                        (ALL,66.23306394)
                    };
            
                    \addplot[
                        dotted,
                        color=green,
                        mark=None, 
                        ultra thick,
                    ] 
                    coordinates {
                        (0,65.7636416)
                        (1,65.7636416)
                        (2,65.7636416)
                        (5,65.7636416)
                        (10,65.7636416)
                        (20,65.7636416)
                        (100,65.7636416)
                        (200,65.7636416)
                        (500,65.7636416)
                        ({1,000},65.7636416)
                        (ALL,65.7636416)
                    };
            
                    \addplot[
                        color=red,
                        mark=*,
                        ultra thick,
                    ]
                    coordinates {
                        (0,57.572312)
                        (1,57.96591997146600)
                        (2,58.02352786064150)
                        (5,58.37371110916130)
                        (10,58.774898)
                        (20,60.217890)
                        (100,61.718253)
                        (200,62.369317)
                        (500,64.538896)
                        ({1,000},66.947786)
                        (ALL,67.659818)
                    };
            
                    \addplot[
                        color=blue,
                        mark=triangle*,
                        ultra thick,
                    ]
                    coordinates {
                        (0,48.315343)
                        (1,52.52095699310300)
                        (2,51.29447698593130)
                        (5,52.39340066909790)
                        (10,58.156070)
                        (20,58.921984)
                        (100,61.457877)
                        (200,62.433885)
                        (500,65.320218)
                        ({1,000},66.710864)
                        (ALL,67.748990)
                    };
                \end{axis}
            \end{tikzpicture}
        \end{adjustbox}
        \caption{Spanish}
    \end{subfigure}
    \hfill
    \begin{subfigure}[b]{0.32\textwidth}
        \centering
        \begin{adjustbox}{width=\textwidth}
            \begin{tikzpicture}
                \begin{axis}[
                    xlabel={{Number of target language examples}},
                    ylabel={\small{F1-score [\%]}},
                    xmin=0, xmax=ALL,
                    symbolic x coords={0, 1, 2, 5, 10, 20, 100, 200, 500, {1,000}, ALL},
                    xtick=data,
                    ymin=37, ymax=69,
                    ytick={40, 45, 50, 55, 60, 65},
                    legend pos=south east,
                    ymajorgrids=true,
                    xmajorgrids=true,
                    grid style={black, dotted},
                    ylabel near ticks,
                    axis line style={black},
                    tick style={black},
                    label style={black},
                    ticklabel style={black},
                    xticklabel style={rotate=45},
                    legend style={font=\footnotesize, text=black},
                    ]
            
                    \addplot[
                        dashed,
                        color=orange,
                        mark=None, 
                        ultra thick,
                    ] 
                    coordinates {
                        (0,58.90996456)
                        (1,58.90996456)
                        (2,58.90996456)
                        (5,58.90996456)
                        (10,58.90996456)
                        (20,58.90996456)
                        (100,58.90996456)
                        (200,58.90996456)
                        (500,58.90996456)
                        ({1,000},58.90996456)
                        (ALL,58.90996456)
                    };
            
                    \addplot[
                        dotted,
                        color=green,
                        mark=None, 
                        ultra thick,
                    ] 
                    coordinates {
                        (0,59.03894424)
                        (1,59.03894424)
                        (2,59.03894424)
                        (5,59.03894424)
                        (10,59.03894424)
                        (20,59.03894424)
                        (100,59.03894424)
                        (200,59.03894424)
                        (500,59.03894424)
                        ({1,000},59.03894424)
                        (ALL,59.03894424)
                    };
            
                    \addplot[
                        color=red,
                        mark=*,
                        ultra thick,
                    ]
                    coordinates {
                        (0,50.38528025)
                        (1,50.44323444366450)
                        (2,50.89148521423340)
                        (5,51.23852491378780)
                        (10,50.97909451)
                        (20,52.53302455)
                        (100,52.73263335)
                        (200,54.94471431)
                        (500,57.70616889)
                        ({1,000},58.27957392)
                        (ALL,59.45787907)
                    };
            
                    \addplot[
                        color=blue,
                        mark=triangle*,
                        ultra thick,
                    ]
                    coordinates {
                        (0,50.38546085)
                        (1,50.04803061485290)
                        (2,49.58138942718500)
                        (5,50.23330986499780)
                        (10,49.7619766)
                        (20,50.59420764)
                        (100,52.40859747)
                        (200,54.24951315)
                        (500,56.93514585)
                        ({1,000},58.87523174)
                        (ALL,60.19658327)
                    };
                \end{axis}
            \end{tikzpicture}
        \end{adjustbox}
        \caption{French}
    \end{subfigure}
    \vfill
    \begin{subfigure}[b]{0.32\textwidth}
        \centering
        \begin{adjustbox}{width=\textwidth}
            \begin{tikzpicture}
                \begin{axis}[
                    xlabel={{Number of target language examples}},
                    ylabel={\small{F1-score [\%]}},
                    xmin=0, xmax=ALL,
                    symbolic x coords={0, 1, 2, 5, 10, 20, 100, 200, 500, {1,000}, ALL},
                    xtick=data,
                    ymin=37, ymax=69,
                    ytick={40, 45, 50, 55, 60, 65},
                    legend pos=south east,
                    ymajorgrids=true,
                    xmajorgrids=true,
                    grid style={black, dotted},
                    ylabel near ticks,
                    axis line style={black},
                    tick style={black},
                    label style={black},
                    ticklabel style={black},
                    xticklabel style={rotate=45},
                    legend style={font=\footnotesize, text=black},
                    ]
            
                    \addplot[
                        dashed,
                        color=orange,
                        mark=None, 
                        ultra thick,
                    ] 
                    coordinates {
                        (0,57.56781697)
                        (1,57.56781697)
                        (2,57.56781697)
                        (5,57.56781697)
                        (10,57.56781697)
                        (20,57.56781697)
                        (100,57.56781697)
                        (200,57.56781697)
                        (500,57.56781697)
                        ({1,000},57.56781697)
                        (ALL,57.56781697)
                    };
            
                    \addplot[
                        dotted,
                        color=green,
                        mark=None, 
                        ultra thick,
                    ] 
                    coordinates {
                        (0,62.86225319)
                        (1,62.86225319)
                        (2,62.86225319)
                        (5,62.86225319)
                        (10,62.86225319)
                        (20,62.86225319)
                        (100,62.86225319)
                        (200,62.86225319)
                        (500,62.86225319)
                        ({1,000},62.86225319)
                        (ALL,62.86225319)
                    };
            
                    \addplot[
                        color=red,
                        mark=*,
                        ultra thick,
                    ]
                    coordinates {
                        (0,50.43097556)
                        (1,49.78069067001340)
                        (2,50.18608748912810)
                        (5,48.97714734077450)
                        (10,50.39399207)
                        (20,52.33710289)
                        (100,54.2673099)
                        (200,55.2574718)
                        (500,56.64525986)
                        ({1,000},58.11224461)
                        (ALL,58.54173541)
                    };
            
                    \addplot[
                        color=blue,
                        mark=triangle*,
                        ultra thick,
                    ]
                    coordinates {
                        (0,47.7152735)
                        (1,48.06140542030330)
                        (2,49.07729804515830)
                        (5,49.70647096633910)
                        (10,51.22304559)
                        (20,52.48954773)
                        (100,54.07661915)
                        (200,57.03153729)
                        (500,60.04134893)
                        ({1,000},62.09627271)
                        (ALL,63.02287459)
                    };
                \end{axis}
            \end{tikzpicture}
        \end{adjustbox}
        \caption{Dutch}
    \end{subfigure}
    \hfill
    \begin{subfigure}[b]{0.32\textwidth}
        \centering
        \begin{adjustbox}{width=\textwidth}
            \begin{tikzpicture}
                \begin{axis}[
                    xlabel={{Number of target language examples}},
                    ylabel={\small{F1-score [\%]}},
                    xmin=0, xmax=ALL,
                    symbolic x coords={0, 1, 2, 5, 10, 20, 100, 200, 500, {1,000}, ALL},
                    xtick=data,
                    ymin=37, ymax=69,
                    ytick={40, 45, 50, 55, 60, 65},
                    legend pos=south east,
                    ymajorgrids=true,
                    xmajorgrids=true,
                    grid style={black, dotted},
                    ylabel near ticks,
                    axis line style={black},
                    tick style={black},
                    label style={black},
                    ticklabel style={black},
                    xticklabel style={rotate=45},
                    legend style={font=\footnotesize, text=black},
                    ]
            
                    \addplot[
                        dashed,
                        color=orange,
                        mark=None, 
                        ultra thick,
                    ] 
                    coordinates {
                        (0,66.378479)
                        (1,66.378479)
                        (2,66.378479)
                        (5,66.378479)
                        (10,66.378479)
                        (20,66.378479)
                        (100,66.378479)
                        (200,66.378479)
                        (500,66.378479)
                        ({1,000},66.378479)
                        (ALL,66.378479)
                    };
            
                    \addplot[
                        dotted,
                        color=green,
                        mark=None, 
                        ultra thick,
                    ] 
                    coordinates {
                        (0,66.96156025)
                        (1,66.96156025)
                        (2,66.96156025)
                        (5,66.96156025)
                        (10,66.96156025)
                        (20,66.96156025)
                        (100,66.96156025)
                        (200,66.96156025)
                        (500,66.96156025)
                        ({1,000},66.96156025)
                        (ALL,66.96156025)
                    };
            
                    \addplot[
                        color=red,
                        mark=*,
                        ultra thick,
                    ]
                    coordinates {
                        (0,54.88150716)
                        (1,55.46613216400140)
                        (2,55.48339128494260)
                        (5,57.45049715042110)
                        (10,57.53723979)
                        (20,58.0497694)
                        (100,58.77814174)
                        (200,61.29332185)
                        (500,63.0670917)
                        ({1,000},65.21175623)
                        (ALL,66.47476792)
                    };
            
                    \addplot[
                        color=blue,
                        mark=triangle*,
                        ultra thick,
                    ]
                    coordinates {
                        (0,48.63176465)
                        (1,52.26499438285830)
                        (2,53.14574599266050)
                        (5,52.32528209686270)
                        (10,53.55931878)
                        (20,57.77267218)
                        (100,58.88597608)
                        (200,60.28836608)
                        (500,63.20851207)
                        ({1,000},65.35504103)
                        (ALL,66.26167417)
                    };
                \end{axis}
            \end{tikzpicture}
        \end{adjustbox}
        \caption{Russian}
    \end{subfigure}
    \hfill
    \begin{subfigure}[b]{0.32\textwidth}
        \centering
        \begin{adjustbox}{width=\textwidth}
            \begin{tikzpicture}
                \begin{axis}[
                    xlabel={{Number of target language examples}},
                    ylabel={\small{F1-score [\%]}},
                    xmin=0, xmax=ALL,
                    symbolic x coords={0, 1, 2, 5, 10, 20, 100, 200, 500, {1,000}, ALL},
                    xtick=data,
                    ymin=37, ymax=69,
                    ytick={40, 45, 50, 55, 60, 65},
                    legend pos=south east,
                    ymajorgrids=true,
                    xmajorgrids=true,
                    grid style={black, dotted},
                    ylabel near ticks,
                    axis line style={black},
                    tick style={black},
                    label style={black},
                    ticklabel style={black},
                    xticklabel style={rotate=45},
                    legend style={font=\footnotesize, text=black},
                    ]
            
                    \addplot[
                        dashed,
                        color=orange,
                        mark=None, 
                        ultra thick,
                    ] 
                    coordinates {
                        (0,53.93941045)
                        (1,53.93941045)
                        (2,53.93941045)
                        (5,53.93941045)
                        (10,53.93941045)
                        (20,53.93941045)
                        (100,53.93941045)
                        (200,53.93941045)
                        (500,53.93941045)
                        ({1,000},53.93941045)
                        (ALL,53.93941045)
                    };
            
                    \addplot[
                        dotted,
                        color=green,
                        mark=None, 
                        ultra thick,
                    ] 
                    coordinates {
                        (0,54.09674525)
                        (1,54.09674525)
                        (2,54.09674525)
                        (5,54.09674525)
                        (10,54.09674525)
                        (20,54.09674525)
                        (100,54.09674525)
                        (200,54.09674525)
                        (500,54.09674525)
                        ({1,000},54.09674525)
                        (ALL,54.09674525)
                    };
            
                    \addplot[
                        color=red,
                        mark=*,
                        ultra thick,
                    ]
                    coordinates {
                        (0,43.77132937)
                        (1,43.89733612537380)
                        (2,44.05258893966670)
                        (5,46.53822779655450)
                        (10,46.26712561)
                        (20,47.35083699)
                        (100,50.08983791)
                        (200,53.44271064)
                        (500,54.95196462)
                        ({1,000},54.9018836)
                        (ALL,58.20449233)
                    };
            
                    \addplot[
                        color=blue,
                        mark=triangle*,
                        ultra thick,
                    ]
                    coordinates {
                        (0,39.10416543)
                        (1,41.97661995887750)
                        (2,43.65054666996000)
                        (5,45.07926881313320)
                        (10,45.28304756)
                        (20,44.80308414)
                        (100,49.70489621)
                        (200,53.8988471)
                        (500,54.26863909)
                        ({1,000},56.51316762)
                        (ALL,56.6433394)
                    };
                \end{axis}
            \end{tikzpicture}
        \end{adjustbox}
        \caption{Turkish}
    \end{subfigure}
    {\ref{mylegend}}
\caption{Effect of adding target language examples on cross-lingual TASD performance with mT5, with and without constrained decoding (CD), compared to monolingual models.}
\label{fig:target_few_shot}
\end{figure}
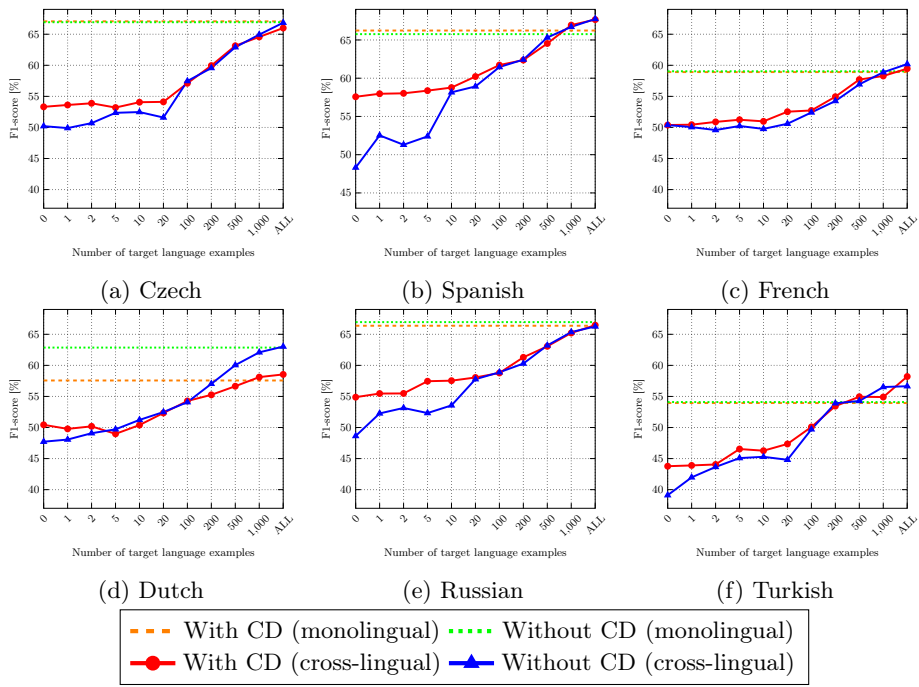

We investigate the impact of increasing the number of target-language examples in the training data for the TASD task, with results shown in Figure~\ref{fig:target_few_shot}. As expected, adding more examples in the target language generally improves performance. For most languages, there is a clear upward trend, with models often approaching or even surpassing monolingual baselines. However, the gains tend to plateau around 1,000 examples, highlighting a practical ceiling given the cost of obtaining high-quality annotations in multiple languages.

As discussed previously, even a small addition of ten examples can yield substantial improvements over the zero-shot setting. Interestingly, this benefit is often more pronounced when constrained decoding is not used. This suggests that a few target-language examples may help compensate for typical generation errors, such as producing aspect terms in the wrong language -- an issue that constrained decoding was designed to mitigate.

With more target-language data, the relative advantage of constrained decoding gradually diminishes. Below 100 examples, constrained decoding can still offer some improvements, but its impact becomes marginal as the number increases. A notable exception is Dutch, where performance with constrained decoding consistently lags behind as more examples are added. In contrast, Dutch models without constrained decoding show marked improvements once the data reaches 200 examples or more.

\subsection{Error Analysis}

We conduct an error analysis to identify the most challenging sentiment prediction elements. We manually examine 100 random test samples from the best-performing mT5 runs -- with and without constrained decoding -- focusing on Czech and Spanish for TASD with up to 100 few-shot examples.

The most frequent errors involve aspect term prediction. As noted in~\cite{icaart25}, the model sometimes outputs aspect terms in the source language instead of the target. This issue, along with typos correction (e.g. \textit{\quotes{se\textbf{vr}ice}} instead of \textit{\quotes{service}}) and hallucinated words, is reduced by constrained decoding and few-shot examples. We also observe incomplete, irrelevant, or missing aspect terms. Error rates tend to decrease with more target-language few-shot examples, aligning with overall performance trends.

Aspect category errors are less common. Rare categories like \textit{\quotes{drinks prices}} are often missed, and similar ones such as \textit{\quotes{restaurant general}} and \textit{\quotes{restaurant miscellaneous}} are frequently confused. Some categories, like \textit{\quotes{food general}}, appear only in one language, which hinders cross-lingual transfer.

Sentiment polarity errors are the least frequent and mainly involve misclassifying the \textit{\quotes{neutral}} class, likely due to label imbalance, since \textit{\quotes{neutral}} is the least frequent class across all datasets.

\section{Conclusion}
This paper investigates the effect of incorporating few-shot target language examples into training data for cross-lingual aspect-based sentiment analysis using sequence-to-sequence models. Across four ABSA tasks, six target languages, and two multilingual models, we show that even a small number of target language examples -- particularly ten -- can lead to significant performance improvements, often rendering techniques like constrained decoding unnecessary. With larger few-shot sets, performance can exceed monolingual baselines, highlighting the strong potential of minimal in-language supervision. These findings offer a practical and cost-effective alternative to zero-shot cross-lingual approaches, especially valuable in low-resource and domain-specific scenarios where obtaining a handful of high-quality annotations is feasible.

\section*{Acknowledgements}
This work has been partly supported by the project R\&D of Technologies for Advanced Digitalization in the Pilsen Metropolitan Area (DigiTech) No. CZ.02.01.01/00/23\_021/0008436. 
Computational resources were provided by the e-INFRA CZ project (ID:90254), supported by the Ministry of Education, Youth and Sports of the Czech Republic.

\bibliographystyle{splncs04}
\bibliography{bibliography}

\end{document}